\documentclass[conference]{IEEEtran}
\usepackage{algorithm}
\usepackage{algpseudocode}
\usepackage{cite}
\usepackage{amsmath,amssymb,amsfonts}
\usepackage{amsthm}
\usepackage{graphicx}
\usepackage{textcomp}
\usepackage{xcolor}

\begin{document}

\title{Addressing Data Heterogeneity in Federated \\ 
Learning of Cox Proportional Hazards Models}

\author{\IEEEauthorblockN{Navid Seidi}
\IEEEauthorblockA{\textit{Dept. of Computer Science} \\
\textit{Missouri S\&T},
Rolla, USA \\
nseidi@mst.edu}
\and
\IEEEauthorblockN{Satyaki Roy}
\IEEEauthorblockA{\textit{Dept. of Mathematical Sciences} \\
\textit{Uni. of Alabama in Huntsville}, USA \\
sr0215@uah.edu}
\and
\IEEEauthorblockN{Sajal K. Das}
\IEEEauthorblockA{\textit{Dept. of Computer Science} \\
\textit{Missouri S\&T},
Rolla, USA \\
sdas@mst.edu}
\and
\IEEEauthorblockN{Ardhendu Tripathy}
\IEEEauthorblockA{\textit{Dept. of Computer Science} \\
\textit{Missouri S\&T},
Rolla, USA \\
astripathy@mst.edu}
}

\maketitle

\begin{abstract}
The diversity in disease profiles and therapeutic approaches between hospitals and health professionals underscores the need for patient-centric personalized strategies in healthcare. 
Alongside this, similarities in disease progression across patients can be utilized to improve prediction models in survival analysis. The need for patient privacy and the utility of prediction models can be simultaneously addressed in the framework of Federated Learning (FL). 
This paper outlines an approach in the domain of federated survival analysis, specifically the Cox Proportional Hazards (CoxPH) model, with a specific focus on mitigating data heterogeneity and elevating model performance.
We present an FL approach that employs feature-based clustering to enhance model accuracy across synthetic datasets and real-world applications, including the Surveillance, Epidemiology, and End Results (SEER) database. 
Furthermore, we consider an \textit{event-based reporting} strategy that provides a dynamic approach to model adaptation by responding to local data changes.
Our experiments show the efficacy of our approach and discuss future directions for a practical application of FL in healthcare.
\end{abstract}


\vspace{-0.05in}
\section{\textbf{Introduction}} \label{sec:Introduction}

Determining the probability of occurrence of specific events is fundamental to biomedical research and clinical studies. \textit{Survival analysis} (SA) is a statistical learning approach for clinical data that uses patient features as inputs and labels indicating event occurrence and observation time. It is crucial in handling \textit{censored} data, which is caused because subjects or patients may not experience the event within the study period or may drop out before the event occurs. Statistical models trained on censored data are used to predict the possibility of the event occurrence for other patients at a specific time. These predictions can be enhanced by incorporating geographic location-based public health features into survival analysis models~\cite{10183765}. The rise of personalized medicine further underscores the need for integrating diverse and high-dimensional data to augment the effectiveness of predictive models in healthcare \cite{masciocchi2022federated, rahman2022fedpseudo}.

Safeguarding patient privacy in healthcare is an important consideration. Modern privacy preserving techniques \cite{tripathy2019privacy, venugopal2022privacy, chen2024generative} can be used to remove sensitive information from datasets without impacting its utility to a large degree. Additionally, health informatics demands strict adherence to protocols ensuring patient privacy~\cite{PRights,hipaa} and data confidentiality. The advent of \textit{federated learning} (FL)~\cite{mcmahan2017communication} enables learning from diverse datasets while mitigating data privacy and ownership concerns \cite{masciocchi2022federated, froelicher2021truly}. It creates robust and collaborative models across multiple institutions without compromising patient-level data privacy, thus facilitating multi-center studies \cite{lu2015webdisco, rahimian2022practical}. Recent studies have focused on harnessing FL to enhance model performance and manage heterogeneous data across healthcare centers, involving optimization and deep learning to tackle concerns of data heterogeneity and privacy~\cite{chowdhury2021review,thwal2021attention,wang2023afei}. However, the substantial volumes of healthcare center data necessitate the exploration of lightweight FL approaches for effective and privacy-preserving solutions. Our research delves into advancing \textit{federated survival analysis} (FSA) by developing and validating lightweight methodologies to address key challenges in model interpretability and data heterogeneity, employing a combination of weighted averaging and event-based reporting mechanisms within the FL framework.

Our paper makes novel contributions to FSA by developing methods that address data heterogeneity and improving model performance. We describe a new method for component-wise weighted averaging, which  improves model accuracy in both synthetic and real-world datasets such as the Surveillance, Epidemiology, and End Results (SEER) \cite{SEER} database. Our paper also studies event-based reporting in federated learning, offering a dynamic approach to model adaptation based on local data changes. Finally, it tackles the challenge of feature space heterogeneity, proposing strategies to manage this complexity effectively in an FL environment. By enhancing model accuracy and interpretability, our contributions are relevant for personalized medicine and healthcare. Since the manual decision-making process at each center is a key determinant of health outcomes, each hospital center is likely to deal with unique pathology and distinct therapeutic approaches. By accommodating these differences as part of personalization, we aim to improve the applicability of FSA.

The paper is organized as follows. Section \ref{sec:Background} reviews the concepts of survival analysis and the role of FL in this domain. Section \ref{sec:methodology} details the proposed FL Framework including data preparation and model development strategies, 
local calculation and parameters broadcasting, and convergence criteria. Section \ref{sec:results} presents experimental results, encompasses component-wise weighted average and event-based Reporting methods, and analyzes the same and heterogeneous feature space. Section \ref{sec:conclusion} concludes the paper.

\section{\textbf{Background and Related Work}} \label{sec:Background}
Survival analysis is a statistical technique for analyzing the time until an event of interest. e.g., death, healing, disease recurrence, etc. 
A commonly used model in survival analysis is the \textit{Cox proportional hazards} (CoxPH) model~\cite{cox1972regression} that estimates the hazard rate, which is the instantaneous rate of occurrence of an event at a point in time, given that no event has occurred till that time, contingent upon various covariates. 
The hazard function of the CoxPH model for an individual, characterized by a set of covariates \(x\), is defined by: 
\begin{equation}
\label{eq:CoxPH_hazard}
    h(t|x) = h_0(t) \exp(\beta^T x),
\end{equation}
\noindent
where \(h_0(t)\) denotes the baseline hazard function describing the hazard rate, serving as a benchmark against which the effects of covariates are measured. The vector \(\beta\) quantifies the influence of a covariate on the hazard rate. These coefficients are a measure of the relative risk associated with the covariates. This model is semiparametric as it does not assume a specific functional form for the baseline hazard function, allowing flexibility with diverse data types.


Federated learning (FL) is increasingly recognized for its potential to develop privacy-preserving methodologies for survival analysis in healthcare, e.g., the Federated Cox Proportional Hazards Model studied in \cite{masciocchi2022federated}.  \cite{archetti2023scaling} proposed federated survival forests adapt FL to heterogeneous datasets. Additionally, the incorporation of geographic location-based public health features into survival analysis has been a subject of growing interest, highlighting the potential of such features in enhancing the predictive accuracy of survival models \cite{10183765}. Current research aims to merge FL with these models, reflecting the shift towards distributed learning approaches in healthcare \cite{lu2015webdisco, rahimian2022practical}. The approaches studied in \cite{rahman2022fedpseudo} jointly balance model integrity with data privacy, aligning with the needs of contemporary healthcare data analysis.

Our methodology addresses the unique challenges of heterogeneous multicentric data in a privacy-conscious FL environment of healthcare settings.

\section{\textbf{Proposed Methodology}}
\label{sec:methodology}

We use the Cox Proportional Hazards (CoxPH) model within an FL framework, addressing data heterogeneity in patient counts and feature spaces across diverse medical centers. 
In our FL adaptation, the CoxPH model is employed across multiple medical centers, each contributing to the collective understanding of survival risks while preserving the privacy and integrity of their data. The federated framework ensures that the model learns from diverse patient data, enhancing the generalizability and robustness of the survival predictions.
\subsection{\textbf{Parameter Estimation in CoxPH Model}}
\label{subsec:beta-calculation-coxph}
The coefficient vector \(\beta\) is estimated by maximizing the partial likelihood over a dataset of patient features and potentially censored times of events. For a dataset of \(r\) individuals, $X = \{x_i | 1\le i \le r\}$, the partial likelihood is defined as:
\begin{equation}\label{eq:likely}
    L(\beta\mid X) = \prod_{i: \delta_i = 1} \left[ \frac{\exp(\beta^T x_i)}{\sum_{j \in R(t_i)} \exp(\beta^T x_j)} \right],
    \vspace{-0.07in}
\end{equation}
where \(x_i\) is the covariate vector for the \(i\)-th individual, \(\delta_i\) is the event indicator (1 if the event occurred, 0 otherwise), \(R(t_i)\) is the set of individuals who have not experienced the event or censoring before time $t_i$ (also called risk set at time \(t_i\)), and \(\beta\) is the coefficient vector. 
We maximize the likelihood using iterative optimization techniques (such as Newton-Raphson), where at each iteration, the \(\beta\) values are updated to increase the partial likelihood. We use the software package \texttt{sksurv}~\cite{sksurv} to learn the value for \(\beta\).
We employ Breslow's method~\cite{breslow1975analysis} for estimating the baseline hazard $h_0(t)$ in \eqref{eq:CoxPH_hazard}. The estimated baseline hazard at each distinct failure time \(t_i\) is given by
\begin{equation}
    \hat{h}_0(t_i) = \frac{d_i}{\sum_{j \in R(t_i)} \exp(\beta^T x_j)},
\end{equation}
where \(d_i\) is the number of events at time \(t_i\).

\subsection{\textbf{Federated Learning Framework}}
\label{subsec:federated-learning-framework}

\textbf{Local Calculations and Beta Estimation.} Every participating center independently estimates its local \(\beta\) coefficients using the CoxPH model \ref{subsec:beta-calculation-coxph} from its partial view of the data:
\begin{equation}
\label{eq:beta_local}
        \beta_{\text{local}, k} = \arg\max L(\beta \mid X_k),
\end{equation}
    where $X_k = \{x_{k, i} : i = 1, 2, \ldots, r_k\}$ denotes the local dataset having $r_k$ patients at the \(k\)-th center.

\textbf{Communication with Global Server.} Each center conveys its \(\beta_{\text{local}, k}\) coefficients to the global server, accompanied by the corresponding feature names and a weighing factor given by the number of rows/patients (\(r_k\)) in its dataset. 


\textbf{Global Beta Aggregation.}
    The global server uses weighted averaging~\cite{konevcny2016federated} to aggregate the local \(\beta\) coefficients from $K$ centers, using the number of patients from each center to ensure a proportionate influence in the global model:
    \begin{equation}
    \label{eq:mg}
        \beta_{\text{global}} = \frac{\sum_{k=1}^{K} r_k \cdot \beta_{\text{local}, k}}{\sum_{k=1}^{K} r_k}.
    \end{equation}

\textbf{Dissemination and Utilization of Global Beta.}
    Upon computing \(\beta_{\text{global}}\), the global server distributes it back to all the centers. Each center updates its local CoxPH model with these global beta coefficients, thereby refining the survival rate predictions for its patients. 
\subsection{\textbf{Heterogeneous Feature Space}}
\label{subsec:heterogeneous-feature-space}

A critical challenge in federated survival analysis, particularly when adapting the Cox Proportional Hazards (CoxPH) is the heterogeneity of feature spaces across centers. The presence and quality of features can vary greatly across centers due to factors such as medical procedures, local regulations, equipment availability, staff expertise, and data collection methodologies~\cite{cottrell2019variation,wiley2022electronic}. This variability in features can impact the effectiveness and applicability of survival analysis models developed using FL. Consider the SEER dataset~\cite{SEER}, where certain features exhibit null or unspecified values in some registries, reflecting the inconsistency in data recording across different geographic locations and healthcare facilities \cite{SEER}. Also, studies on Electronic Health Records (EHR) data, which are a primary source for survival analysis, have highlighted the variability in data quality and feature availability across centers \cite{cottrell2019variation, wiley2022electronic}, impeding the development of a unified, robust federated model for survival analysis.

To address this issue, we propose \textit{Naive Global Models Parameters Averaging} and \textit{Component-wise Weighted Average Approach}. These approaches are designed to navigate the complexities of varying feature spaces, ensuring that the FL model remains resilient and applicable across diverse healthcare settings. 


\subsection{\textbf{Naive Global Models Parameters Averaging}}
\label{subsec:naive-global-models-parameters-averaging}

This method aggregates the CoxPH model parameters across centers only for those features that are present across all centers (FedAvg~\cite{mcmahan2017communication}).
Let $F_k$ denote the features present in the dataset at center~$k$. The algorithm begins by identifying a subset of features common across all participating centers:
\begin{equation}
\label{eq:common}
    F_{\text{common}} = \bigcap_{k=1}^{K} F_{k}.
\end{equation}
For the common features, the CoxPH parameters are computed using a weighted average. So, equation~\ref{eq:mg} is updated as:
\begin{equation}
\label{eq:global}
    \beta_{\text{global}} = \frac{\sum_{k=1}^{K} r_k \cdot \beta_{\text{local}, k}(F_{\text{common}})}{\sum_{k=1}^{K} r_k}
\end{equation}
Here, \(\beta_{\text{local}, k}(F_{\text{common}})\) are the $\beta$ coefficients for the common features at center \(k\). 
Once \(\beta_{\text{global}}\) is calculated, it is distributed to all centers. Each center then updates its CoxPH model with these global coefficients for the common features. (Refer to Algorithm~\ref{alg:naive-g} for pseudo-code). 
\begin{algorithm}
\begin{algorithmic}[1]
\caption{Naive Global Models Parameters Averaging}
\label{alg:naive-g}
\State \textbf{Input:} Datasets at $K$ centers $\{X_1, X_2, \ldots, X_K\}$
\State \textbf{Output:} Global CoxPH model $\beta_{\text{global}}$

\State Calculate $F_{\text{common}}$ using \eqref{eq:common} 
\Comment{Common features}

\For{each center $k$ locally in parallel} \Comment{Local models}
    \State Calculate $\beta_{\text{local}, k}(X_k)$ using \eqref{eq:beta_local}
    \State $r_k \gets \text{Number of rows in } X_k$
\EndFor

\For{each feature $f$ in $F_{\text{common}}$} \Comment{At global server}
    \State $\beta_{\text{global}}(f) \gets \frac{\sum_{k=1}^{K} r_k \cdot \beta_{\text{local}, k}(f)}{\sum_{k=1}^{K} r_k}$
\EndFor

\For{each center $k$ locally in parallel} \Comment{Download}
    \State Update local CoxPH model at center $k$ with $\beta_{\text{global}}$
\EndFor
\State \Return $\beta_{\text{global}}$
\end{algorithmic}
\end{algorithm}

\subsection{\textbf{FL with Feature Presence Clustering}}
\label{subsec:component-wise-weighted-average-clustering}

To effectively address the feature space heterogeneity in Federated Survival Analysis (FSA), we propose a clustering approach based on the presence of features across centers. This method aligns with the frameworks suggested in \cite{ghosh2022efficient, nafea2022proportional, sattler2020clustered}, but introduces specific adaptations for survival analysis, particularly leveraging the unique properties of the Cox Proportional Hazards (CoxPH) model. 
In this approach, each center initially constructs a binary vector representing the presence (1) or absence (0) of each feature in its dataset, forming a feature presence bit pattern, $\mathbf{B}_k$ for center $k$. This binary pattern is shared during the federated learning rounds, enabling a clustering process that groups centers with similar feature availability.
The feature presence vector for center $k$ with feature set $F_k$ is defined as:
\begin{equation}
    \mathbf{B}_k = [b_{k,1}, b_{k,2}, \dots, b_{k,p}]
    \label{eq:bit-feature}
\end{equation}
where $b_{k,i} = 1$ if the $i$-th feature is present in center $k$, and $0$ otherwise. This method is similar to \cite{ghosh2022efficient}, but adapted to the heterogeneous characteristics typical in healthcare datasets.

Next, we employ a clustering algorithm, specifically K-Means clustering, to group centers based on their feature presence vectors, minimizing the within-cluster sum of squares, a common objective in data clustering \cite{ghosh2022efficient}:
\begin{equation}
    \min_{C_1, C_2, \dots, C_c} \sum_{i=1}^c \sum_{k \in C_i} \|\mathbf{B}_k - \mu_i\|_{Hamming}
    \label{eq:k-means-obj}
\end{equation}

\noindent Here, $C_i$ denotes the set of centers in cluster $i$, $c$ is the number of clusters, and $\mu_i$ is the mean bit pattern of cluster $i$. 

This clustering facilitates more tailored and effective aggregation of survival models, as centers with similar data structures can share statistical strength without the noise introduced by vastly different feature sets. Post clustering, for each cluster, we compute a weighted average of the CoxPH model's $\beta$ coefficients for features common within the cluster. This is akin to the component-wise weighted averaging proposed in \cite{nafea2022proportional}, but here, it is confined within identified clusters:
\begin{equation}
    \label{eq:beta_global,i}
    \beta_{\text{global},i} = \frac{\sum_{k \in C_i} r_k \beta_{\text{local},k}}{\sum_{k \in C_i} r_k}
\end{equation}
where $r_k$ is the number of patients in center $k$, and $\beta_{\text{local},k}$ are the local CoxPH coefficients from center $k$. 

\begin{figure}[h]
    \centering
    \includegraphics[width=0.9\linewidth]{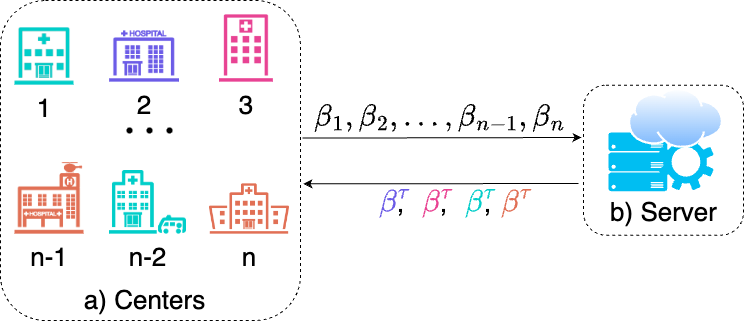}
    \caption{FL Framework illustration with Feature Presence Clustering. (a) Centers compute local model parameters $\beta_1,  \ldots, \beta_n$ and send them to (b) the central Server to aggregate them and update global model parameters $\beta^*$.}
    \label{fig:fl-clustering}
\end{figure}

Figure~\ref{fig:fl-clustering} illustrates the communication and parameter exchange process in our federated learning setup. It shows how centers share their local values with the server. The server then clusters the centers based on feature presence, calculates the global \(\beta\) for each cluster, and sends them back to the centers.
Algorithm~\ref{alg:feature-presence-clustering} gives the pseudo-code of the procedure.
\begin{algorithm}
\begin{algorithmic}[1]
\caption{FL with Feature Presence Clustering}
\label{alg:feature-presence-clustering}
\State \textbf{Input:} Datasets at $K$ centers $\{X_1, X_2, \ldots, X_K\}$, number of clusters $c$
\State \textbf{Output:} CoxPH models $\{\beta_{\text{global}, 1}, \beta_{\text{global}, 2}, \ldots, \beta_{\text{global}, c}\}$
\State Each center~$k$ uploads its feature presence vector $B_k$ \eqref{eq:bit-feature}.
\State Global server obtains cluster assignments using \eqref{eq:k-means-obj}.
\For{each center $k$ locally in parallel} \Comment{Local models}
    \State Calculate $\beta_{\text{local}, k}(X_k)$ using \eqref{eq:beta_local}
    \State $r_k \gets \text{Number of rows in } X_k$
\EndFor
\For{each cluster $C_i$} \Comment{At global server}
    \State Calculate $\beta_{\text{global}, i}$ using \eqref{eq:beta_global,i}.
\EndFor
\For{each center $k$ locally in parallel} \Comment{Download}
    \State Update local model at $k$ with $\beta_{\text{global}, i}$ if $k \in C_i$.
\EndFor
\State \Return $\beta_{\text{global}, 1}, \beta_{\text{global}, 2}, \ldots, \beta_{\text{global}, c}$
\end{algorithmic}
\end{algorithm}

\subsection{\textbf{Convergence Guarantee for Algorithm~\ref{alg:feature-presence-clustering}}}
\label{subsec:convergence-guarantee}
As described in~\eqref{subsec:beta-calculation-coxph}, the Cox Proportional Hazards (CoxPH) model employs a loss function derived from the partial likelihood of survival times, specifically structured for survival analysis. The partial likelihood function is shown in~\ref{eq:likely}. The loss function utilized for parameter estimation is the negative log of this partial likelihood:
\begin{equation*}
    \mathcal{L}(\beta) = -\sum_{i: \delta_i = 1} \left[ \beta^T x_i - \log\left(\sum_{j \in R(t_i)} \exp(\beta^T x_j)\right) \right]
\end{equation*}
This function captures individual contributions and the collective risk dynamics at each event time, which is crucial for understanding survival outcomes under the CoxPH model.

\theoremstyle{plain}
\newtheorem{theorem}{Theorem}
\newtheorem{assumption}{Assumption}

\begin{assumption}[L-Smoothness]
The gradient of the loss function is Lipschitz continuous, ensuring that there exists a constant \( L > 0 \) such that 
\begin{equation}
    \label{eq:ls0}
  \mathcal{L}(\beta_1) \leq \mathcal{L}(\beta_2) +  (\beta_1 - \beta_2)^T  \nabla \mathcal{L}(\beta_2)+ \frac{L}{2} \| \beta_1 - \beta_2 \|^2,
  \vspace{-0.05in}
\end{equation}
\begin{equation}
    \label{eq:ls1}
\|\nabla \mathcal{L}(\beta_1) - \nabla \mathcal{L}(\beta_2)\| \leq L\|\beta_1-\beta_2\|
\vspace{-0.05in}
\end{equation}
for any parameters \( \beta_1, \beta_2 \).
\end{assumption}

\begin{assumption}[Strong Convexity]
The loss function is strongly convex, i.e., there exists a constant \( \mu > 0 \) such that 
\begin{equation}
    \label{eq:sc0}
\mathcal{L}(\beta_1) \geq \mathcal{L}(\beta_2) + (\beta_1 - \beta_2)^T  \nabla \mathcal{L}(\beta_2) + \frac{\mu}{2}\|\beta_1 - \beta_2\|^2,
\vspace{-0.05in}
\end{equation}
\end{assumption}
for any parameters \( \beta_1, \beta_2 \).
\begin{assumption}[Proper clusters]
The cluster members revealed by \eqref{eq:k-means-obj} are such that $\sum_{k \in C_i}\mathcal{L}_k$ satisfies Assumptions 1 and 2 for all $i\in \{1, 2, \ldots, c\}$.
\end{assumption}

\begin{theorem}[Convergence of Algorithm~\ref{alg:feature-presence-clustering}]
Given the assumptions of smoothness, strong convexity, and proper clusters, Algorithm~\ref{alg:feature-presence-clustering} converges to the optimal solution \(\beta_{\text{global}, i}^*\) of the Cox Proportional Hazards model for each cluster~$i$ if the learning rate satisfies $\eta < \mu/L^2$. The convergence behavior is characterized by:
\[
\left\|\beta^{(t+1)}-\beta_{\text{global}, i}^*\right\|^2 \leq\left(1-\eta \mu(1-\eta \frac{L^2}{\mu})\right)^T \left\|\beta^{(0)}-\beta_{\text{global}, i}^*\right\|^2,
\]
where $\beta^{(t)}$ denotes the global parameter for cluster $C_i$ at iteration $t$ of FL and $T$ is the total number of FL iterations.
\end{theorem}

\begin{proof}
Under the proper cluster assumption, the reasoning for one cluster of centers applies to all the other clusters. Hence we drop the cluster subscript index and denote the global loss function for a particular cluster by \(\mathcal{L}(\beta)\) and its optimal parameter by \(\beta^*\). 
   Using a superscript to denote the $t$th federated learning iteration, each participating center \( k \) updates its local parameters \(\beta_k\) as:
   \[
   \beta_k^{(t+1)} = \beta^{(t)} - \eta \nabla \mathcal{L}_k(\beta^{(t)}).
   \]
   The server aggregates the updates from centers in cluster~$C_i$ as per \eqref{eq:beta_global,i} to update the global parameter \(\beta^{(t+1)}\):
   \begin{equation}\label{eq:cluster update}
   \beta^{(t+1)} = \frac{\sum_{k \in C_i} r_k \beta^{(t+1)}_{k}}{\sum_{k \in C_i} r_k}
   = \beta^{(t)} - \frac{\sum_{k \in C_i} r_k \nabla \mathcal{L}_k(\beta^{(t)})}{\sum_{k \in C_i} r_k}
   \end{equation}
   where \( r_k \) is the number of data points at center \( k \) and $\mathcal{L}(\beta^{(t)}) = \frac{\sum_{k \in C_i} r_k \nabla \mathcal{L}_k(\beta^{(t)})}{\sum_{k \in C_i} r_k}$.
   Considering $\beta_1=\beta^{(t)}$ and $\beta_2=\beta^*$, since $\nabla \mathcal{L}(\beta^{*}) = 0$, from the strong convexity assumption (\eqref{eq:sc0}), we have:
\begin{equation}
\label{eq:sc}
       \mathcal{L}(\beta^{(t)}) - \mathcal{L}(\beta^*) \geq \frac{\mu}{2} \|\beta^{(t)} - \beta^*\|^2
       \vspace{-0.05in}
\end{equation}
\begin{equation}
\label{eq:sc2}
        \|\beta^{(t)} - \beta^*\|^2 \leq \frac{2}{\mu}(\mathcal{L}(\beta^{(t)}) - \mathcal{L}(\beta^*) )
        \vspace{-0.05in}
\end{equation}
From \eqref{eq:cluster update} we get that:
\[
    \|\beta^{(t+1)}-\beta^{*}\|^2 = \|\beta^{(t)}-\beta^{*}-\eta\nabla\mathcal{L}(\beta^{(t)})\|^2.
\]
Next, using the expansion equation for the norm $(\|a - b\|^2 = \|a\|^2 - 2a^T b + \|b\|^2)$ where \(a = \beta^{(t)} - \beta^*\) and \(b = \eta \nabla \mathcal{L}(\beta^{(t)})\), we have:
\begin{multline}
       \|\beta^{(t+1)}-\beta^{*}\|^2 = \|\beta^{(t)}-\beta^*\|^2\\-2 \eta((\beta^{(t)}-\beta^*)^T \nabla \mathcal{L}(\beta^{(t)}))+\eta^2\|\nabla \mathcal{L}(\beta^{(t)})\|^2 
       \vspace{-0.05in}
\end{multline}
\begin{multline}
\label{eq:exp2}
      (\beta^{(t)}-\beta^*)^T \nabla \mathcal{L}(\beta^{(t)}) = (\|\beta^{(t)}-\beta^*\|^2\\-\|\beta^{(t+1)}-\beta^{*}\|^2+\eta^2\|\nabla \mathcal{L}(\beta^{(t)})\|^2)/  2 \eta
      \vspace{-0.05in}
\end{multline}
Setting $\beta_1=\beta^*$ and $\beta_2=\beta^{(t)}$ in strong convexity we have:
\begin{equation}
\label{eq:sc3}
\left(\beta^{(t)}-\beta^*\right)^T \nabla \mathcal{L}\left(\beta^{(t)}\right) \geq \mathcal{L}\left(\beta^{(t)}\right)-\mathcal{L}\left(\beta^*\right)+\frac{\mu}{2}\left\|\beta^{(t)}-\beta^*\right\|^2
\end{equation}
Substituting right-hand side of~\eqref{eq:exp2} in left-hand side of~\eqref{eq:sc3} and multiplying with $2\eta$:
\begin{multline}
        \|\beta^{(t)}-\beta^*\|^2-\|\beta^{t+1}-\beta^{*}\|^2+\eta^2\|\nabla \mathcal{L}(\beta^{(t)})\|^2 \geq \\ 2\eta (\mathcal{L}\left(\beta^{(t)}\right)-\mathcal{L}\left(\beta^*\right)+\frac{\mu}{2}\left\|\beta^{(t)}-\beta^*\right\|^2)
\end{multline}
\begin{multline}
        -\|\beta^{t+1}-\beta^{*}\|^2 \geq -\|\beta^{(t)}-\beta^*\|^2 -\eta^2\|\nabla \mathcal{L}(\beta^{(t)})\|^2\\+2\eta (\mathcal{L}\left(\beta^{(t)}\right) -\mathcal{L}\left(\beta^*\right)+\frac{\mu}{2}\left\|\beta^{(t)}-\beta^*\right\|^2)
        \vspace{-0.05in}
\end{multline}
\begin{multline}
\label{eq:exp3}
        \|\beta^{t+1}-\beta^{*}\|^2 \leq \|\beta^{(t)}-\beta^*\|^2 +\eta^2\|\nabla \mathcal{L}(\beta^{(t)})\|^2\\-2\eta (\mathcal{L}\left(\beta^{(t)}\right) -\mathcal{L}\left(\beta^*\right)+\frac{\mu}{2}\left\|\beta^{(t)}-\beta^*\right\|^2)
        \vspace{-0.05in}
\end{multline}

\noindent Next, squaring both sides of \eqref{eq:ls1} with $\beta_1 = \beta^{(t)}$, $\beta_2=\beta^{*}$, and $\nabla\mathcal{L}(\beta^*)=0$, we have:

\[
\|\nabla \mathcal{L}(\beta^{(t)})\|^2 \leq L^2\|\beta^{(t)}-\beta^*\|^2
\]
Combining the above with~\eqref{eq:sc2} and simplifying, we have:
\[
\|\nabla \mathcal{L}(\beta^{(t)})\|^2 \leq \frac{2L^2}{\mu} \left( \mathcal{L}(\beta^{(t)}) - \mathcal{L}(\beta^*) \right)
\]
Now,~\eqref{eq:exp3} can be updated as:
\begin{multline}
   \|\beta^{t+1}-\beta^{*}\|^2 \leq \left\|\beta^{(t)}-\beta^*\right\|^2+2\eta^2 \frac{L^2}{\mu}\left(\mathcal{L}\left(\beta^{(t)}\right)-\mathcal{L}\left(\beta^*\right)\right)\\-2 \eta\left(\mathcal{L}\left(\beta^{(t)}\right)-\mathcal{L}\left(\beta^*\right)+\frac{\mu}{2}\left\|\beta^{(t)}-\beta^*\right\|^2\right)
\end{multline}
Combining like terms:
\begin{multline}
\|\beta^{t+1}-\beta^{*}\|^2 \leq\left\|\beta^{(t)}-\beta^*\right\|^2-2 \eta\left(\mathcal{L}\left(\beta^{(t)}\right)-\mathcal{L}\left(\beta^*\right)\right)\\+\eta \mu\left\|\beta^{(t)}-\beta^*\right\|^2+2 \eta^2 \frac{L^2}{\mu}\left(\mathcal{L}\left(\beta^{(t)}\right)-\mathcal{L}\left(\beta^*\right)\right)
\end{multline}
Factoring out $(\mathcal{L}(\beta^{t})-\mathcal{L}(\beta^{*}))$ and substituting~\eqref{eq:sc}, we have:
\begin{multline}
\left\|\beta^{(t+1)}-\beta^*\right\|^2 \leq\left\|\beta^{(t)}-\beta^*\right\|^2\\-2 \eta(1-\eta \frac{L^2}{\mu})\left(\frac{\mu}{2}\left\|\beta^{(t)}-\beta^*\right\|^2\right)+\eta \mu\left\|\beta^{(t)}-\beta^*\right\|^2
\end{multline}
Combining $\|\beta^{(t)}-\beta^{*}\|^2$, we have:
\begin{equation}
\left\|\beta^{(t+1)}-\beta^*\right\|^2 \leq(1-\eta \mu(1-\eta \frac{L^2}{\mu}))\left\|\beta^{(t)}-\beta^*\right\|^2
\end{equation}
For convergence, we need:
\[
0 < 1 - \eta \mu (1 - \eta \frac{L^2}{\mu}) < 1
\Leftrightarrow
\eta \mu (1 - \eta \frac{L^2}{\mu}) > 0,
\]
which is true under the following conditions
\[
0 < \eta < \frac{\mu}{L^2}, \quad \mu > 0, \quad L > 0.
\]

\end{proof}
\vspace{-0.2in}

\subsection{\textbf{Event-Based Reporting Method}}
\label{subsec:event-based-reporting-method}

This module aims to minimize communication overheads by allowing the centers to communicate with the global server selectively, transmitting information only when specific events occur. The reporting is guided by a performance threshold parameter at the global server, denoted by \(\epsilon\), which is used to determine the efficacy of each center's local model in contributing to the FL process.

\subsubsection{Threshold Definition and Distribution}
Initially, the global server sets and communicates a threshold \(\epsilon\) to all centers after the first federated learning (FL) round, denoting the minimum model accuracy improvement for subsequent rounds. Centers then assess their performance improvements against this benchmark, where \(\epsilon\) is the given threshold value.

\subsubsection{Local Model Accuracy Evaluation}
After computing local model parameters in each subsequent round, every center calculates its current model accuracy, using the Concordance Index as the metric. The center then compares this accuracy with that of the previous round. The Concordance Index at round \(t\) for center \(k\) is denoted as \(CI_{k}^{(t)}\).

\subsubsection{Decision for Participation in Federated Learning}
Each center determines its eligibility for participation in the next round of FL based on the change in its C-Index. If the change in accuracy falls below the threshold \(\epsilon\), the center opts out of sending its local parameters for the next round. This decision is formulated as:
\begin{equation*}
    \parbox{3cm}{Participation by center $k$ in FL round $t$}
    = 
    \begin{cases} 
    \text{Yes,} & \text{if } CI_{k}^{(t)} - CI_{k}^{(t-1)} \geq \epsilon, \\
    \text{No,} & \text{otherwise.}
    \end{cases}
\end{equation*}

\vspace{0.02in}
\section{\textbf{Experimental Results}}
\label{sec:results}

For the evaluations of the proposed approach, we used both simulated and real-world datasets to validate the efficacy of the proposed FL approach. The code for this study is publicly available on the project's GitHub webpage~\cite{STM}. 

\begin{figure*}[h!]
\centering
\vspace{-0.2in}
\includegraphics[width=1.0\textwidth, height = 7cm]{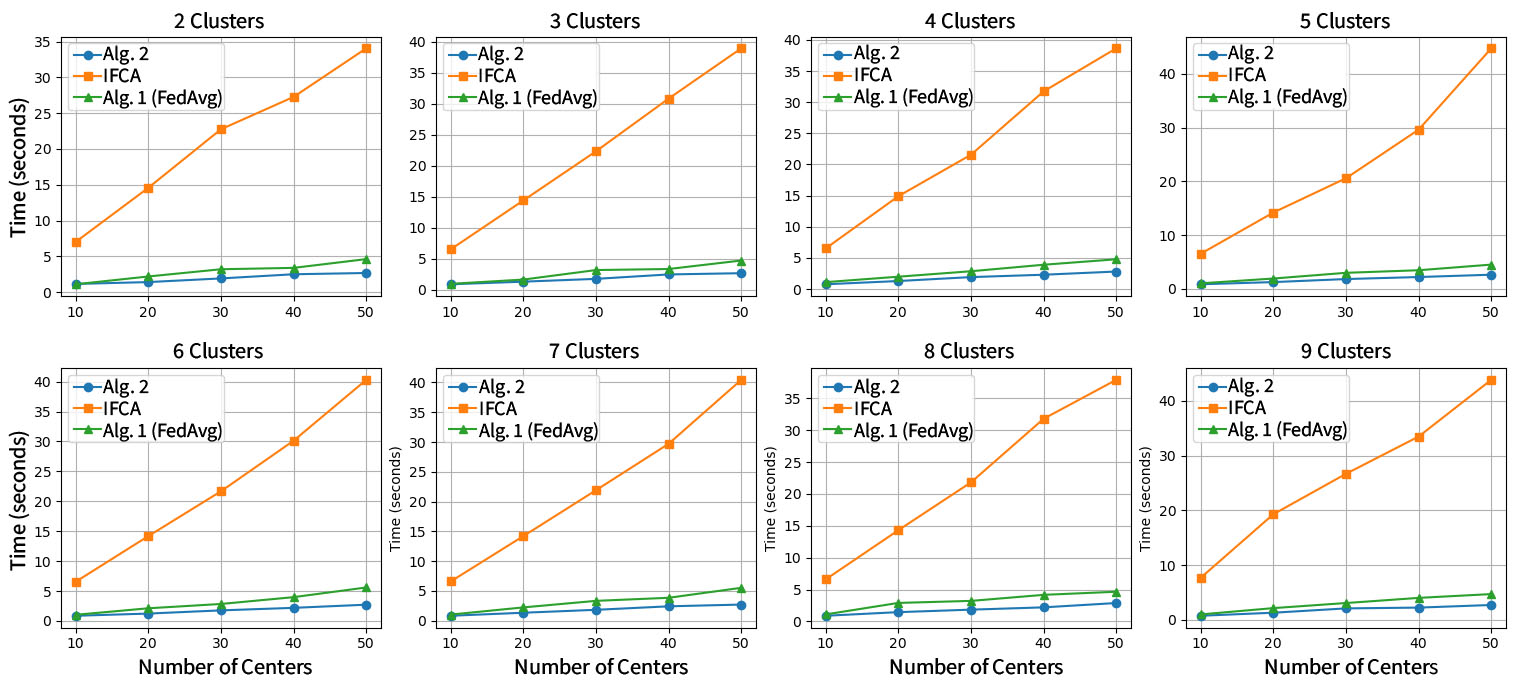}
\vspace{-0.2in}
\caption{Running times of the different algorithms (Alg. 1: FedAvg, Alg. 2: proposed algorithm, IFCA) for various numbers of clusters and centers.}
\label{img:running_time.png}
\end{figure*}

\subsection{Simulation Studies}
Our simulation studies, conducted across 50 centers with patient counts ($r$) randomly assigned between 900 and 1100, demonstrated enhanced performance over traditional FL approaches. Synthetic datasets were generated to include a total of 100 features per patient: 11 common features across all centers and a random selection from the remaining 89 to model real-world data variability. Features for each patient, \( \mathbf{x}^{(i)} \), were sampled from a normal distribution \( \mathbf{x}^{(i)} \sim \mathcal{N}(0, I_p/p) \), where $p$ is the number of features, ensuring \( \mathbf{E}\left[\lVert \mathbf{x}^{(i)} \rVert^2 \right] = 1 \). Here, \( I_p \) represents the identity matrix of size \( p \times p \) implying that each feature is independently and identically distributed with a variance of \( \frac{1}{p} \). We modeled actual event times \(\tau_i \) from a Cox proportional hazards model with a constant baseline hazard \( \lambda_0 \), and incorporated right censoring from a uniform distribution \( U(0, \Psi^{-1}(\frac{1}{2})) \) where $\Psi(u) \equiv \mathbb{P}[\tau<u]=1-e^{-u \exp \left(\beta^T \boldsymbol{x}_i\right)}$~\cite{andreux2020federated}. The observed times, determined as the minimum of censoring and actual event times, simulate a FSA environment with decentralized and heterogeneous data. We used the Iterative Federated Clustering Algorithm (IFCA) presented in \cite{ghosh2022efficient} as the baseline for our clustering approach.

\subsubsection{Component-Wise Weighted Average with Feature Presence Clustering}
Table~\ref{tab:clustering} shows the improvements in local C-Index score in each center for Component-Wise Weighted Average Approach in conjunction with Feature Presence Clustering to handle feature space heterogeneity in FSA. 

The consistently higher positive rates in Algorithm~\ref{alg:feature-presence-clustering} along with the statistically significant p-values, underscore the efficacy of our clustering technique. These results demonstrate the substantial benefits of our method in enhancing predictive accuracy in FSA across various clustering configurations.

\begin{table}[h!] 
\centering
\caption{Number of centers that improve C-Index using FL in a simulated experiment with 50 centers}
\label{tab:clustering}
\begin{tabular}{|c|c|c|c|c|c|}
\hline
\textbf{\# Clusters} & \textbf{Alg. 1} & \textbf{IFCA} & \textbf{Alg. 2} & \textbf{T-Statistic} & \textbf{P-Value} \\
\hline
2 &  44 & 46 & \textbf{48} & 2.56 & 0.011 \\
\hline
3 &  39 & 42 & \textbf{45} & 1.75 & 0.085 \\
\hline
4 &  37 & 46 & \textbf{46} & 2.89 & 0.004 \\
\hline
5 &  37 & 43 & \textbf{44} & 2.10 & 0.037 \\
\hline
6 &  38 & 45 & \textbf{47} & 3.22 & 0.001 \\
\hline
7 &  37 & 38 & \textbf{40} & 1.98 & 0.048 \\
\hline
8 &  37 & 42 & \textbf{43} & 2.60 & 0.009 \\
\hline
9 &  37 & 43 & \textbf{44} & 2.63 & 0.019 \\
\hline
\end{tabular}
\end{table}

\subsubsection{Running Times of Algorithms}
Figure~\ref{img:running_time.png} presents the running times of the different algorithms (Alg. 1: FedAvg, Alg. 2: proposed algorithm, IFCA) for varying numbers of clusters and centers. The results indicate that our method (Alg. 2) consistently demonstrates lower running times compared to IFCA across all clustering configurations. This significant reduction in running time, particularly noticeable as the number of clusters increases, highlights the efficiency of our approach. While FedAvg (Alg. 1) also shows lower running times, it does not provide the same level of predictive accuracy improvements as our proposed method, as evidenced in Table~\ref{tab:clustering}.


\subsubsection{Event-Based Reporting Method}
In our experiment, which applied an event-based reporting method in FL (as detailed in Subsection~\ref{subsec:event-based-reporting-method}), we executed five rounds of learning. In each round, we adjusted the dataset size by \textbf{adding} or \textbf{removing} a variable number of rows/patients: $\mathbf{0 < \text{small} < 50}$, $\mathbf{0 < \text{medium} < 100}$, and $\mathbf{0 < \text{large} < 200}$.

\begin{figure}[h!]
\centering
\includegraphics[width=0.50\textwidth]{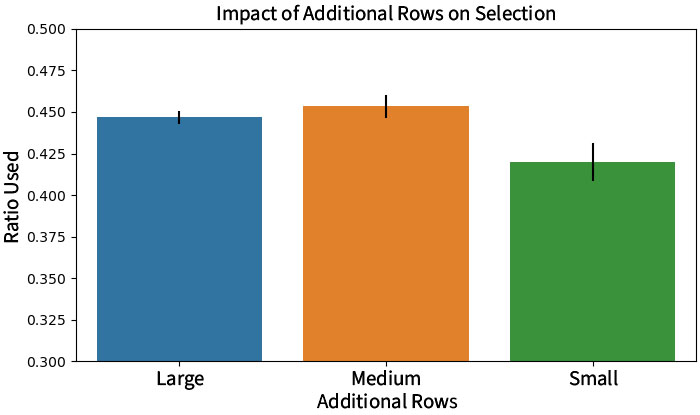}
\caption{Effect of dataset size on the selection frequency across learning rounds.}
\label{img:ratio_bar_std}
\end{figure}

This method, driven by a performance threshold parameter (\(\epsilon = 1 \times 10^{-5}\)), was designed to minimize communication overheads by allowing selective transmission of data from centers to the server. We chose this value for \(\epsilon\) because it demonstrated the best performance in our setup. Note that this value can be learned as a hyperparameter and fine-tuned based on the specific applications and datasets. Adjusting \(\epsilon\) to the characteristics of the data can optimize the balance between communication overhead and model performance.

Notably, datasets with no additional rows were consistently not selected for participation. This process was repeated 25 times to obtain standard error bars, reflecting the precision of averages in the CoxPH analysis. Figure~\ref{img:ratio_bar_std} shows the average 'Selected' status ratio for additional patient data groups.

\subsection{Real-World Data Application}
We applied our approach to the SEER dataset, representing breast cancer cases in ten states between 2000 and 2017. Treating each state as a federated node, we captured the inherent heterogeneity of the data. This approach not only improved accuracy but also underscored the value of data diversity. Table~\ref{table:seer_cindex_comparison} exhibits that the federated average consistently improves upon local calculations. Moreover, the weighted average by rows (number of patients) further surpasses the naive average by the number of centers, highlighting the effectiveness of our FL approach in enhancing predictive accuracy.

\begin{table}[htbp]
    \caption{C-Index Comparison across Different  Methods in SEER Dataset}
    \vspace{-0.1in}
    \begin{center}
    \begin{tabular}{|c|c|c|c|}
    \hline
    \textbf{Dataset} & \textbf{Local Calcs.} & \textbf{Wtd. Avg. Centers} & \textbf{Wtd. Avg. Rows} \\
    \hline
    KY & \textbf{0.72374} & 0.70591 & 0.71813 \\
    \hline
    LA & 0.71758 & 0.72768 & \textbf{0.73594} \\
    \hline
    CT & 0.50000 & 0.73390 & \textbf{0.73720} \\
    \hline
    UT & 0.50000 & 0.72369 & \textbf{0.72433} \\
    \hline
    GA & 0.71963 & 0.72038 & \textbf{0.72432} \\
    \hline
    HI & \textbf{0.77666} & 0.73105 & 0.76109 \\
    \hline
    NJ & 0.71805 & 0.71914 & \textbf{0.72831} \\
    \hline
    NM & 0.72504 & 0.71592 & \textbf{0.72740} \\
    \hline
    IA & 0.78413 & 0.79920 & \textbf{0.81941} \\
    \hline
    CA & 0.72317 & 0.72499 & \textbf{0.73437} \\
    \hline

    \end{tabular}
    \label{table:seer_cindex_comparison}
    \end{center}
        \vspace{-0.1in}
\end{table}


\section{\textbf{Conclusion}}
\label{sec:conclusion}
This work addresses data heterogeneity in patient counts and feature space for federated survival analysis. 
Our approach leverages weighted averaging techniques and event-based reporting mechanisms to achieve significant improvements in model accuracy and efficiency. Specifically, our component-wise weighted average approach and feature presence clustering effectively managed data heterogeneity, while the event-based reporting minimized communication overhead. 
The potential impacts of our work include enhanced predictive accuracy and personalized healthcare strategies, aligning with smart health initiatives like those in \cite{seidi2017vid} and \cite{seidi2019pv}. 
Future work can explore hyperparameter tuning of the performance threshold (\(\epsilon\)) to optimize communication overhead. 
\bibliographystyle{IEEEtran}
\bibliography{main}
\end{document}